# What Makes it Difficult to Understand a Scientific Literature?

Mengyun Cao[#*1], Jiao Tian[#*2], Dezhi Cheng[#*3], Jin Liu[&4], Xiaoping Sun[*5]

[#]*University of Chinese Academy of Sciences*
*Beijing, China*
[1]caomengyun@kg.ict.ac.cn
[2]tianjiao@kg.ict.ac.cn
[3]chengdezhi@kg.ict.ac.cn

[*]*Knowledge Grid Group*
*Key Lab of Intelligent Information Processing*
*Institute of Computing Technology*
*Chinese Academy of Sciences*
[5]sunxiaoping@ict.ac.cn

[&]*Wuhan University*
[4]jinliu@whu.edu.cn

[*]*Abstract*—In the artificial intelligence area, one of the ultimate goals is to make computers understand human language and offer assistance. In order to achieve this ideal, researchers of computer science have put forward a lot of models and algorithms attempting at enabling the machine to analyze and process human natural language on different levels of semantics. Although recent progress in this field offers much hope, we still have to ask whether current research can provide assistance that people really desire in reading and comprehension. To this end, we conducted a reading comprehension test on two scientific papers which are written in different styles. We use the semantic link models to analyze the understanding obstacles that people will face in the process of reading and figure out what makes it difficult for human to understand a scientific literature. Through such analysis, we summarized some characteristics and problems which are reflected by people with different levels of knowledge on the comprehension of difficult science and technology literature, which can be modelled in semantic link network. We believe that these characteristics and problems will help us re-examine the existing machine models and are helpful in the designing of new one.

## I. INTRODUCTION

The term Artificial Intelligence (AI) was coined by John McCarthy in 1955 [25] and gradually developed into a formal discipline. This domain is usually defined as the science and engineering of making machines, especially intelligent computer programs, conduct tasks that require intelligence when done by humans [1]. Natural language processing (NLP) is among the central goals of AI research [2] and becomes an attractive research field. Some of tasks in NLP are used to solve syntax/grammar analysis tasks, such as word segmentation, co-reference resolution, named entity recognition, Part-of-Speech (PoS) tagging, etc. Some of the tasks such as automatic summarization, question answering, and machine translation have high level real-world applications [3] for assisting people in reading, information retrieval and mining. The performance of models aiming at addressing these high level tasks of NLP is still far from satisfactory. For example, the state-of-the-art results on automatic summarization are not quiet readable yet [6]. Almost all question-answering systems can only handle questions that are either based on single-relation or factual issues with some simple inference and their performance can only achieve 50% on average using this domain's evaluation criteria.

To achieve a better solution, one way is to tackle a high level task into many sub tasks and apply different methods to solve them in more effective and efficient way. Recently, statistics machine learning methods such as topic models [20] and deep neural network (NN) models [16][17][18] have achieved significant progress on many sub NLP tasks [20]. For example, the F1-score of word segmentation for Chinese novels can reach more than 90% by using a common noun entities mining method [4], a new model for PoS tagging can achieve more than 90% accuracy on different domains [5]. Teaching machine to read and comprehend is even possible [8]. These advances are deemed as a big step toward our ideal.

If we can make machine to read and comprehend text like human, one would be able to make more intelligent task possible, like to teach machine large amount of knowledge. But before we dedicate ourselves to design new models for making machine reading and comprehension possible, shall we rethink what is the help that people really desire or how difficult and what difficulties in that task.

To more concretely feel difficulties in making machine reading and understanding text, we can evaluate how difficult for a human to read and comprehend text. Since most people feel no difficult in comprehending daily reading task such as reading news, we pay our attention to scientific literature comprehension. That is, what makes it difficult for students or researchers to understand an academic paper? In order to answer this question, we organized an experiment about the comprehension of human on scientific literature reading.

We conducted this test by letting participants read academic papers that they may not understand well. It should be noted that we assume that there is no problem in the articles and readers' understanding process is to rebuild the thought of the author inside their own mind. That is, the content in the articles, such as the structure of papers and the conclusion drowned by the author, are all reasonable. Two papers written in different styles and period were selected as our test material. Six people with different levels of knowledge background on computer science were invited to offer their questions when reading test papers.

We intended to probe the comprehension impediments that they encountered through their questions, and anatomize the reasons causing these impediments. After the analytical steps, we draw several conclusions about the characteristics and problems on the comprehension of such difficult science and technology literature by people who have different levels of knowledge. We argue that these characteristics and problems will facilitate the inspection of the existing works on NLP, and will provide some insightful guide for the future research. There are in fact already many psychological works on science text comprehension properties [21]. Their main purpose is to improve the quality of tutoring and teach. We conduct this work mainly from an angle of computer science and our main target is to investigate how we can leverage computer models to do the understanding task.

## II. SEMANTIC AND KNOWLEDGE MODEL

Before we introduce the main experiment and its results, we first introduce how to define the problem and concepts and how to model the problem in a computer science way, rather than in a psychological way. The key is to model the semantics, the knowledge and its relationships with texts or scientific papers. That is, when we say that we can understand a scientific paper, we mean that we can setup a mapping from the text to the semantic and knowledge information in reader's brain and this mapping can match the author's understanding in a certain acceptable degree. Of course, exact mapping is impossible. To understand is at least to be able to set up such a mapping and such a mapping can be accepted by many readers in a common sense. To model this argument, we adopt a semantic link network model [14] to describe the semantics and knowledge in the text of a scientific paper.

### A. Example

A semantic link network is a network consisting of entity names and their relationships. An entity is a string representing a real world object and concept. A relationship is a common accepted relationship such as class-instance relationship, causality relationship, belong-to relationship, negation relationship, etc.. Then the semantic of a string of a word, a sentence or a paragraph, is defined as a mapping from this string to the predefined semantic link network. The knowledge of such a string is the mapped node and its related nodes in the semantic link network. The constraints of semantic link network are not as strict as what has been coined in the Semantic Web languages [19], which gives

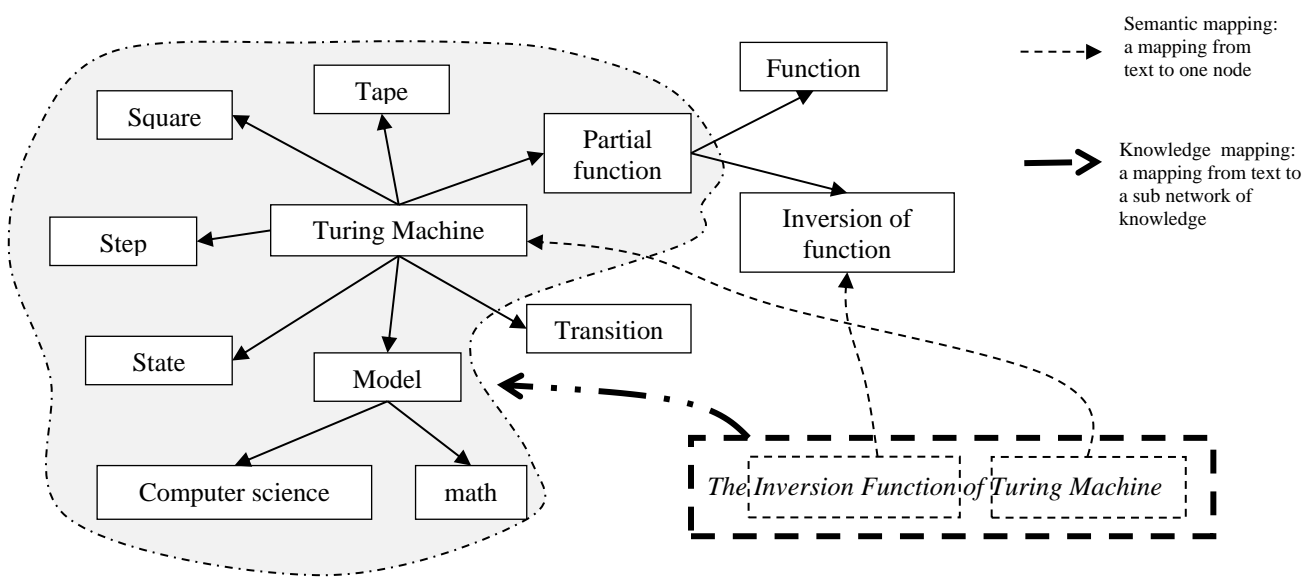

Fig 1. A simple semantic link network for describing a Turing machine model and a mapping from a sentence to the semantic link network

semantic link network a flexible modeling capability.

Fig 1. gives such an example of using semantic link network for describing a Turing machine model and a semantic mapping and a knowledge mapping from a sentence "*The inversion function of Turing Machine*". As shown in the figure, the shaded area on the network is a piece of knowledge of about the Turing machine model. The edges among concepts can be inclusion relationship, equivalence relationship and sub class relationship. The dotted arrow line represented semantic mapping, which is to map an object to a node in the network. The bold dotted arrow line is to map the whole sentence to the knowledge piece in the shaded area. The shade area does not cover the partial function or the transition concept. It is to show this mapping is not complete. It is just an incomplete mapping or even incorrect mapping.

But note that here it does not mean that we use the semantic link network to describe every piece of knowledge in author's mind. Or we do not mean that knowledge is only the semantic link network instances. Rather, we use the semantic link network to model the knowledge of authors in some scale or sense such that the semantic link network can be a knowledge representation at certain level or accuracy or coverage. Or in another word, if we can find the author or a specialist, we can let the author use the semantic link network tool to describe his knowledge in a certain level of details about his paper and we deem this semantic link network as the knowledge representation of the paper of the author or a teacher or an authority. We assume such a network can be detailed and extended by authors or people. And when a reader read the paper, he or she can also use the semantic link network tool to describe his knowledge derived from the paper. In this way, we can set up a computable comparing platform to see what they know about this paper.

### B. Semantic link network based knowledge modelling

So from this example, if we say that a reader can understand the sentence "The *inversion function of Turing machine*", we mean that they can setup such a kind of mapping from text to a semantic link network that describes the related knowledge about the sentence. Of course, a semantic link network about one thing is different for different people with different knowledge background and understanding. We assume that there is such a "correct" semantic link network inside author's understanding. And a 'correct' understanding by a reader is to setup such an approximation to the correct semantic link network and correct mapping from text to it.

A sentence or a text may be mapped to different knowledge in different context. But in fact, a context of a sentence $S$ of a paper $P$ is the knowledge of a set $C$ of strings larger than that original string. $P$ here is the set of any text combination of texts in a paper. So its mapping could be different from the $S$. To clearly show how this can model the reading process. We formally define the related concepts as below:

(1) $P$: a paper.

(2) $s \in P$: a text or a set of text from $P$.

(3) $A=<E,V>$: A semantic link network for describing knowledge of a given people $A$.

(4) $B \subseteq A$ means that a semantic link network $B$ is a sub network of $A$.

(5) A semantic mapping from $s$ to the knowledge is defined as:

$s(A):s \rightarrow v$, where $v \in V$ is a node in the semantic link network $A$.

(6) A knowledge mapping from a text $s \in P$ to knowledge $A(s)$ by a given people $A$ when reading a sentence $s$ can be defined as:

$A(s):s \rightarrow A(s)$, where $A$ is the current background knowledge of reader $r$, *i.e.,* a set of semantic link network having being built before reading P. $A(s)$ is the semantic link network either belong to $A$ or a new semantic link network that is new to $A$.

Also note that a semantic mapping can be deemed as a knowledge mapping: $s(A) = A(s)$ when $A=< E =\varnothing, V=\{v\}>$, i.e. mapping $s$ to a semantic network $A$ with only one node and no edge.

But simply union of $w(A)$ for all $w \subseteq s$ does not construct $A(s)$ because there is no edge in $w(A)$.

(7) $A(P):\{A(s):s \rightarrow A(s)|s \in P\}$ is a collection of semantic link networks derived from text of paper P.

(8) $A$ can be updated by $A(s)$ using a simple graph union operation $A \Leftarrow A \cup A(s)$, which means that a semantic link network $A$ is extended by merging nodes and links from $A(s)$ as well as adding new edges among $A$ and $A(s)$. If $A(s) \subseteq A$, then, $A \cup A(s)=A$. Similarly, $A(s) \Leftarrow A(s) \cup A(p)$ means that the semantic link network derived from $s$ is extended by another semantic link network derived from $p$ and in this case and the newly one can be different from the original one.

$A(p) = A(p) \cup A(s)$ may not equal to the $A(s) = A(s) \cup A(p)$ because new edges are added when applying $A(p) \Leftarrow A(p) \cup A(s)$, which may be different from applying $A(s) \Leftarrow A(s) \cup A(p)$.

(9) $B(s) \approx A(s)$ means that two semantic link networks are similar or the knowledge of two people on $s$ are similar. The similarity can be defined in terms of graph similarity considering both nodes and edges labels as well as their topology or can be directly specified using a semantic link "*equivalent*" or "*similar to*"

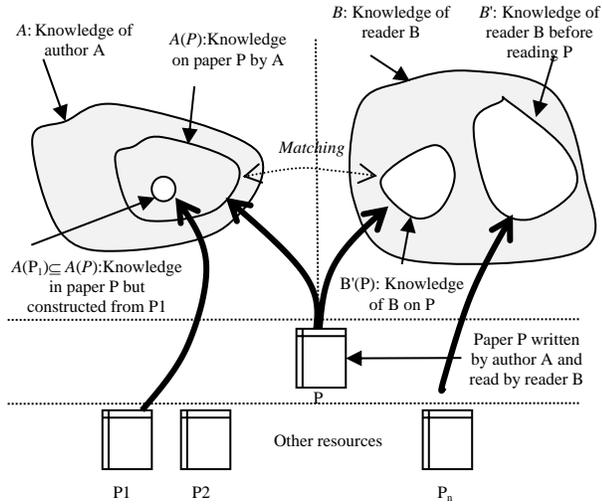

Fig 2. Knowledge derived from paper P by author A and reader B

One possible confusing point is about the abstraction of a semantic link network. For example, one semantic link network $T$ may contain only one node $v=$ "Turing Machine" and another one $A$ is like what in Fig. 1. One may argue that $B(s) \approx A(s)$ where $s =$ "Turing Machine". But in fact, from the aspect of graph information, $T$ is quite different from $A$. We take it as the different knowledge because without any other information we cannot deduce that $T$ is A. One possible of such information is that a direct semantic link "*equivalent*" is added from $T$ to $A$. When adding such an equivalent link between $T$ and $A$, it can be deemed as an updating operation on $T(s)$ by $A(s)$. That is, $T(s) \Leftarrow T(s) \cup A(s)$ adds a link "*equivalent*" between corresponding node in $T$ and in $A$.

## C. Reading and Comprehension modelling

Then, we can now use the above modeling tools to model a reading and comprehension case. We use an example in Fig. 2 to show how this work.

In Fig. 2, an author $A$ has a paper $P$ that is read by a reader $B$. So, here $A$ is used to represent the semantic link network knowledge of the author along with his paper. We also use $A(P)$ to represent the semantic link network derived from P by the author $A$. And there is a semantic link network $A(P_1) \subseteq A(P)$ and $A(P_1)$ is derived from $P_1$ rather than $P$. So, in fact, an extension is applied as $A(P) \Leftarrow A(P) \cup (P_1)$. Then, the process can be modelled in sequence as following:

(1) $A$ is built.
(2) $A(P_1)$ is built
(3) $A(P) \Leftarrow A(P) \cup A (P_1)$ is built. Note that the merging is actually happen during writing each word or sentence by $A$.

$B'$ represents the semantic link network knowledge of the reader $B$ before reading paper $P$. After reading $P$, his semantic link knowledge can be extended by: $B \Leftarrow B' \cup B'(P)$. And then, by applying the updating extension $B(P) \Leftarrow B \cup B'(P)$, we can have the final semantic link network as the knowledge model of $B$ after reading $P$.

But we can see that if the reader $B$ has not accessed the paper $P_1$, then, he may not be able to set up such a knowledge mapping that $B(P) \approx A(P)$.

## III. EXPERIMENT

We conduct a small experiment involving six readers and two scientific papers. Table 1 shows the basic information of six participants majoring in computer science area and their roles in the task. It is also noted that they are all not native English speakers. We divided participants into two groups $G_1$ and $G_2$. There are 4 people in $G_1$ and their questions are used as material for classification and analysis to get conclusions. The questions of $G_2$ are used as the validation material for our conclusions.

One of our test material (referred to as $P_1$) is a classical paper named *"The Inversion of Functions defined by Turing Machines"*. This article was written by John McCarthy in 1956 [7]. It is difficult for most people to understand in three aspects: first, it involves very fundamental ideas on Turing machine, computational complexity and formal theory; second, some of its concepts are too old to maintain its original meaning in current; in addition, there are many inferences and functions that the author considered obviously but cannot be easily proved by readers. Another (referred to as $P_2$) is a newly published paper named *"Teaching machines to read and comprehend"*. This article introduces some deep NN and classical methods for answering Cloze form queries [8], and that may difficult to understand for people who have no contact with these methods.

TABLE I EXPERIMENTER INFORMATION

| Group | Participant ID | Computer Science Degree | Roles |
|---|---|---|---|
| G1 | A | college graduate | ask questions |
|  | B | master candidate | ask questions |
|  | C | master candidate | ask questions |
|  | D | doctor | ask & answer |
| G2 | E | master candidate | ask questions |
|  | F | PhD candidate | ask questions |

TABLE II CLASSIFY THE QUESTIONS ABOUT $P_1$

| | $P_1$ | | | | |
|---|---|---|---|---|---|
| | type0 | type1 | type2 | type3 | Synthesis |
| $G_1$ | 3 | 10 | 37 | 31 | 9 |
| $G_2$ | 0 | 6 | 12 | 5 | 3 |
| A | 3 | 4 | 25 | 24 | 10 |
| B | 0 | 8 | 18 | 15 | 2 |
| C | 0 | 9 | 23 | 9 | 1 |
| D | 0 | 0 | 2 | 0 | 0 |
| E | 0 | 1 | 4 | 3 | 3 |
| F | 0 | 6 | 10 | 4 | 3 |

TABLE IIII CLASSIFY THE QUESTIONS ABOUT $P_2$

| | $P_2$ | | | | |
|---|---|---|---|---|---|
| | type0 | type1 | type2 | type3 | Synthesis |
| $G_1$ | 1 | 12 | 36 | 27 | 7 |
| $G_2$ | 0 | 11 | 20 | 11 | 11 |
| A | 1 | 4 | 18 | 17 | 9 |
| B | 0 | 4 | 11 | 15 | 2 |

| | | | | |
|---|---|---|---|---|
| C | 0 | 9 | 20 | 12 | 7 |
| D | 0 | 0 | 2 | 0 | 0 |
| E | 0 | 11 | 14 | 9 | 10 |
| F | 0 | 0 | 8 | 1 | 2 |

*A. Experiment Processes*

Our experiment is conducted in two days (at least 6 hour a day) without letting junior participants to do much preparation work before. The first day, we handed out the electronic and print version of $P_1$ to all the participants. We let participants read it paragraph by paragraph without any auxiliary material such as dictionary, and meanwhile write down their questions as well as the gist of each paragraph. Then, D singled out some questions to answer. We recorded all the questions and answers from D. The second day, we do the same thing with $P_2$.

We removed repeated questions for the same test paper on the same group. At last, for $P_1$, we collected a total of 90 questions asked by $G_1$ and 26 by $G_2$; for $P_2$, there are 83 questions asked by $G_1$ and 53 by $G_2$. Finally, by analyzing materials at hand, we tried to classify all the questions and summarized several characteristics in the comprehension processes, and showed them in the following sections.

*B. Analysis and Classification*

First, we tried to classify those questions to see what properties are there. These questions can be divided into four types according to their inherent characteristic (see Table II and Table II):

  *1) Type 0: Language problems:*
Language problems represent the questions caused by the grammar, or particular expression of the author. This type contains the least number of questions because the English level of our participants can fit the requirement of parsing the papers. For example:
- Whether the word of *"to"* is missing in the third sentence of paragraph 19? Because we always use the phrase *"from ... to ..."*.

  *2) Type 1: Lack of the semantic mapping from the notations or words to the concepts behind the word:*
Readers did not notice that this notation links to the real concept. For example:
- A does not known the word *"enumerative"* in the first sentence of paragraph 5.
- C does not known the word *"homogeneous"* and *"isotropic"* in the first sentence of paragraph 8.

  *3) Type 2: lack of links to the knowledge outside the paper and reader's current knowledge:*
The knowledge outside of the paper represents the meaning of concepts, logical relationships among concepts, derivation of theorem and formulas, etc., that are mentioned in the particular context of the paper and are not directly provided in the paper but we should have known to comprehend the paper. In another words, the knowledge could not be learned in this articles but are still required in understanding processes.

For example, article $P_1$ does not explain anything about Turing Machine. So, the structure and the operation mechanism of Turning Machine such as *"tape"*, *"square"* and *"internal states"*, which are basics in Turing Machine model, are contained within this paper's outside knowledge.

Following questions are raised by A in $P_1$:
- What are the specific definitions of *"steps"* in the third, the fourth sentence of paragraph 1, the last sentence of paragraph 3 and the first two sentences of paragraph 6?
- Why *"for any Turing machine there is another one which does k steps of the original machine in one step"*? This sentence is in the second sentence of paragraph 6.
- Why *"a machine with Q internal states and S symbols should be considered as making about 1/2logQS elementary steps per step of computation"*? This sentence is in the last sentence of paragraph 8.

  *4) Type 3: lack of semantic links inside the paper:*
The answers of this type of questions could be founded out through the description in the articles, but participants cannot find them probably because they have not yet read the relevant sentences in the following sections, or they had some incomprehensible parts in the earlier parts.

Several questions of this type in $P_1$ are listed here for example:
- What is the meaning of *"well-defined problems"* mentioned in the first sentence of paragraph 1? Actually the interpretation of this notation is given in the next three sentences.
- In the fourth sentence of paragraph 7, reader A cannot find the corresponding parts in transform function of *"but"*?
- Which is the corresponding part of "return to the question" mentioned in the first sentence of paragraph 10?

  *5) Synthesis questions:*
The questions of this type are caused by two or more questions listed in the previous type. That is, for example, a question that is raised not only because the language problems but also the lack of links to the knowledge outside of the test papers.

Take the questions raised by $G_1$ about $P_2$ for example:
- What is meaning of paragraph 21? A, B, C all asked this question because they not only have no idea about *Deep LSTM Reader ( both lack of outside knowledge of $P_2$ and the incomprehension about description in the paragraph above )*, but also don't know what the symbols in the formulas refer to.
- Why the author said that *"There is no significant advantage in this"* in the fifth sentence of paragraph 16? The asker cannot understand the reason that the author explains followed the mention.

*C. Semantic link network modelling*

From Table II and III, we can see that the major types of questions are related to the knowledge inside and outside paper. We can use the modeling method in section II to describe these problems. Fig. 3 shows what the knowledge outside the paper is and what knowledge inside the paper is.

Assuming that the paper contains a sentence *S*: "*A well-defined problem is a problem that has Turing Machine tester to validate its solution*", let *R* is reader and *A* is the author, then:

*1) Semantic mapping is missing.*

Type 1 questions can be modeled as $w(R)=\varnothing$ or $w(R) \neq w(A)$ for a word or a phrase $w \in S$. For example, in Fig.3 there are three semantic mappings that link from the words and phrases of sentence to the nodes in this knowledge piece.

*2) Knowledge constructed inside the paper*

There is one semantic link network $R(S)$ that can be constructed from the sentence *S* in the paper (marked in the right shaded area of Fig. 3). This network $R(S)$ is deemed as the knowledge inside the paper *P* because $s \in P$ of reader *R*. If this knowledge cannot be setup by reader *R*, the problem related to it is of type 3.

*3) Knowledge outside the paper and reader knowledge*

There is a semantic link network $A(w)$ of the author *A* about the word $w$ ="*Turing Machine*" which can be detailed in Fig.1. Here we use a shaded area in the left corner to represent such a network. The knowledge is outside the paper because from sentences in the paper reader *R* can NOT setup such a knowledge piece, or a semantic link network to represent the model of Turing machine that is similar to $A(w)$. Setting up $R(w)$ such that $R(w) \approx A(w)$ is possible only when the reader know the $P_0$ before reading because $A(w)= A(w) \cup A(P_0)$ and $P_0$ is other resource. More formally, we can define this problem as:

$\neg \exists s$ such that when $R(w) \Leftarrow R \cup R(w) \cup R(s)$, we have $R(w) \approx A(w)$ and $A(w) \Leftarrow A(w) \cup A(P_0)$ with $s \in P$.

That is, there is no text *s* in paper *P* such that $R(w)$ can be extended to approximate $A(w)$.

So, basically, type 1 problem is about the semantic mapping setup problem. Type 2 problem is about the knowledge mapping to the knowledge outside paper and Type 3 problem is related to the knowledge mapping construction to the inside knowledge. How about type 0 problem? Type0 problem is about the semantic mapping and knowledge mapping from sentences to the language knowledge.

*4) Modelling understanding*

To understand sentence *S* in the figure, of course, *R* need to approximate $A(S)$, which in turn requires to setup $R(w)$ in a correct way. Before setting up $R(w)$, there is already an R(S) which can be derived from *S* using current knowledge *R*. But this $R(S)$ is far from $A(S)$. If *R* obtains $A(w)$ by reading some other materials $P_0$, then, it can be described as :

$R(w) \Leftarrow R(w) \cup R(P_0)$ and then $R(w) \approx A(w)$.

Then, by applying another merging operation:

$R(S) \Leftarrow R(S) \cup R(w)$. Then, we finally can have $R(S) \approx A(S)$ because $R(S) \Leftarrow R(S) \cup R(w)$ and $A(S) \Leftarrow A(S) \cup A(w)$.

### D. Characteristics and problems

Further, we tried to analyse characteristics of those questions raised by readers to see why they may raise such questions. We have found that there are 10 characteristics, or say, problems, that are reflected by these questions.

*1) The key questions that hinder people from understanding the article could not be asked by themselves sometime :*

For example, reader A had a question *"What are the logical relations among 'not defined', 'exist', 'existence', and 'not exist'? Are they not in conflict with each other?"* when A reads the paragraph 2 of $P_1$.

In fact it is because A does not know *"partial function"* and *"$m^{th}$ Turing machine"*, and the relationship between function *'g(m, r)'* and *'$f_m(g(m, r))$'*, which lead A to ask that question. And *A* has set up a wrong mapping for *"partial function"*.

*2) The major understanding obstacles that people*

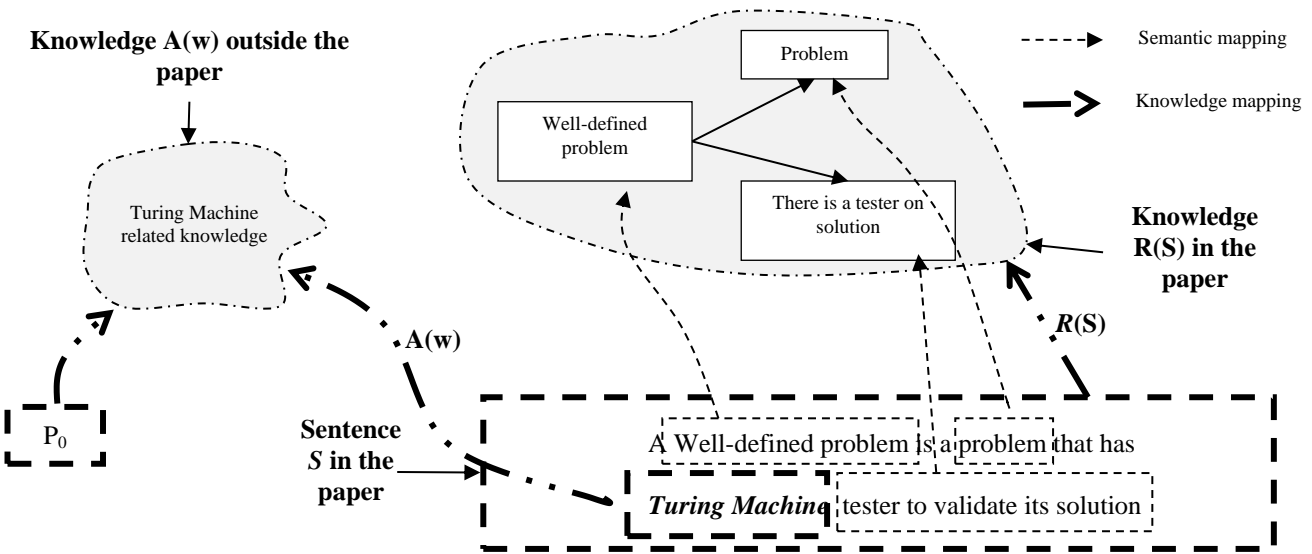

Fig 3. A semantic link network that is derived from the sentences of paper and it is also related to a knowledge piece outside the paper

*concern are mainly focused on type 2 and type 3:*

This conclusion can be directly drawn for the total number of questions in all types. The questions in these two categories are the majority of our question lists for all participants.

Here we assumed that people who may read scientific literature all have a good command of English. As for type 1 questions, they all believed that problems caused by the mismatch about notation and concepts can be solved easily by a dictionary.

*3) The barrier of understanding caused by the lack of links to the knowledge outsid the paper and the reader's knowlege will seriously prevent people from comprehending the paper they are reading:*

First, a lot of questions in type 4 contain the concepts that have been asked in type 2. For example, a question of $P_1$ in type 2 is *"what's the meaning of 'well-defined problems' in first two sentences of paragraph 1?"*, while in $P_1$'s type 3 there is a question about the *"logical relationship among 'problem', 'solution' and 'test' in paragraph 1"*.

Second, as shown in the first characteristic, the key to answer many questions depends on the complete understanding of knowledge outside the papers. People may not discover the importance of some concepts or think they have already armed with that background knowledge, which will finally leads to either wrong understanding or incomprehensible concepts in other parts of the paper.

*4) Some questions of type 3 may also be raised by leaping thinking in articles and readers cannot keep up with the authors' thought:*

For example, in the $P_1$, paragraphs from 7 to 9 discuss a computational complexity in another way. Both B and C have learned related concepts before, but they paid all of their attention on the Turing machine, partial function and the inversion of functions and raised a lot of questions because they did not realize that these paragraphs is about computational complexity theory.

*5) People can't reproduce some proof of theorems and reasoning of functions not only because the lack of background knowledge but also related to their individual factors such as capability and training experience:*

For example, D has enough background to understand the major parts of two test papers but he still cannot prove an argument in the $P_1$: *"a machine with Q internal states and S symbols should be considered as making about 1/2 log QS elementary steps per step of computation"*, which is mentioned in the last sentence of paragraph 8.

*6) The obstacles of comprehension caused by the limitation of outside knowledge could not be overcame by using the internal information of the paper; That is, information from outside is needed:*

This is obvious according to the definition of the knowledge outside the paper. There are the knowledge that cannot be learned within this article but still be need in understanding processes.

*7) People's comprehension of the main ideas weakens dramatically with the increasing of points they don't understand as well as with the accumulation of their bad mood:*

Reader A, B and C could hardly summarize the main ideas of the paragraphs after the third page of $P_1$. E even gave up reading $P_1$ after paragraph 6. As for $P_2$, reader A, B, C and E all jumped from paragraph 21 to 26 because there are many things they don't know for deep NN. And they all feel agitated because there are numerous points they could not understand in the paper.

*8) Most of the time, we still cannot understand the article even if excellent work on syntax/grammar parsing has been done:*

In our experiment, grammar or language problem of type 0 is rare. So, solving type 0 problem may not help much to solve type 1 and 2 problems. Of course, basic syntactic parsing is necessary for understanding basic meaning of sentences, but we argue that the works of syntax parsing can't help understanding content when the understanding requires a certain level of semantics or knowledge that is far above the basic semantics of words.

*9) Some research on semantics could help people understand articles:*

In linguistics, semantics is a study mainly focusing on the meaning and relationship inherent at the levels of words, phrases, sentences, and larger units of discourse [11]. Many tasks in NLP need to concern semantic, such as automatic summarization, co-reference resolution, question answering, etc. As we argued in the introduction, most of them are still far from satisfaction. Although these tools can help some in identifying word and relationships, we argue that to make machine reading and comprehending, we need to pay attentions to the problems of type 2, 3 and 4 we have modelled.

*10) The function of guide is significant:*

D spent a month reading $P_1$ until he could make a comprehension, but A, B and C only take one day to reach a certain degree of understanding under the guidance of D. So the guidance is important.

From this view, if we intend to make a machine simulate human to understand scientific literature and then help people read, it is more plausible that we teach the machine to understand at first than let it study on its own from scratch. There are many methods we can refer in the teaching process among people, such as exemplification, graphical method, searching for the key problems, etc..

## IV. IMPLICATIONS

Although this experiment is conducted in a very small scale with only six participants and two papers involved, we still argue that the results can be representative, especially for those who are reading difficult scientific literatures that they are not so familiar with. Moreover, what we have done in this experiment at least show three instructive points that could be referred when we design artificial intelligence models for letting machine to understand text:

*1) Syntax semantics and content semantics*

A language expression has two levels of semantics. One is syntax semantics that is used to understand a basic, topic-irrelevant semantics of the expression. In our work they are related to type 0 problems. But capturing basic semantics of an expression may not help much for understanding content semantics of text. How to deal with this gap may be fundamental in making machine understanding natural language text.

*2) the knowledge outside the papers.*

To better process text, outside knowledge is important or even indispensable. How to incorporate explicit outside knowledge into machine learning process is an important research issue. In this view, it may be worth thinking how to design the method for machines to express and save knowledge, and how to design the algorithm to learning new knowledge using their own knowledge.

*3) High-level guidance.*

Guidance is an efficient way to increase the efficiency of understanding [23]. How to make such explicit guidance on machine learning process deserves further study. At present, the popular way to "teach" a machine is to input rules defined by people, or by a supervised learning model with large benchmark data. But we still cannot understand how human comprehend articles until now. So we cannot completely define the rules for understanding and teach it to machine.

*4) Semantic link network modelling.*

We have used the semantic link network model to analyze four types of questions and we found that the key to make the reader understanding is to help them setup semantic link networks either from outside knowledge sources or from the texts in the paper and then help them setting up mappings from texts to those semantic link networks. So this can help us derive the future solution to make the machine read and comprehend the text. That is, make machine be able to set up semantic link networks and then setting up mappings from texts to those semantic link networks automatically [15][22][24].

## V. CONCLUSIONS

In this work, we conducted a human reading comprehension experiment where we let six people read two computer science papers which are difficult to understand for many people, and let them write down their questions in the comprehension process. From the experiment, we have summarized ten characteristics and problems that people reflected in reading difficult scientific literature by trying to classify and analyze the question they listed using a semantic network model. And those characteristics and problems show that even for people, it is still hard to comprehend such complicated papers completely. For machine, it is much more difficult to conduct such kinds of task. But the results also provide some implications, that is, we should consider how to deal with syntax semantics and content semantics, and how to incorporate explicit knowledge into machine learning process and how to make explicit guidance on machine learning process.

## ACKNOWLEDGMENT

Thanks to Wei Li and Pengshan Cai for their supporting and thanks to Prof. Zhiwei Xu for his suggestions in this work. Research Supported by the Open Project Funding of CAS Key Lab of Intelligent Information Processing, Institute of Computing Technology, Chinese Academy of Sciences (No. IIP2014-2).This work was also partially supported by National Science Foundation of China (No.61075074 and No.61070183).